\documentclass[letterpaper, 10 pt, conference]{ieeeconf}  

\IEEEoverridecommandlockouts                              

\usepackage{graphicx} 
\usepackage{amsmath} 
\usepackage{amssymb}  
\usepackage{multirow}
\usepackage{makecell}

\title{\LARGE \bf
FEDORA: A Flying Event Dataset fOr Reactive behAvior}

\author{Amogh Joshi, Wachirawit Ponghiran, Adarsh Kosta, Manish Nagaraj and Kaushik Roy\\Purdue University, West Lafayette, IN 47907, USA\\$\{joshi157, wponghir, akosta, mnagara, kaushik\}@purdue.edu$
\thanks{This work was supported by IARPA MicroE4AI.}
}

\begin{document}

\maketitle
\thispagestyle{empty}
\pagestyle{empty}

\begin{abstract}

The ability of resource-constrained biological systems such as fruitflies to perform complex and high-speed maneuvers in cluttered environments has been one of the prime sources of inspiration for developing vision-based autonomous systems.
To emulate this capability, the perception pipeline of such systems must integrate information cues from tasks including optical flow and depth estimation, object detection and tracking, and segmentation, among others. However, the conventional approach of employing slow, synchronous inputs from standard frame-based cameras constrains these perception capabilities, particularly during high-speed maneuvers. Recently, event-based sensors have emerged as low latency and low energy alternatives to standard frame-based cameras for capturing high-speed motion, effectively speeding up perception and hence navigation. For coherence, all the perception tasks must be trained on the same input data. However, present-day datasets are curated mainly for a single or a handful of tasks and are limited in the rate of the provided ground truths.
To address these limitations, we present Flying Event Dataset fOr Reactive behAviour (FEDORA) - a fully synthetic dataset for perception tasks, with raw data from frame-based cameras, event-based cameras, and Inertial Measurement Units (IMU), along with ground truths for depth, pose, and optical flow at a rate much higher than existing datasets.

\end{abstract}

\section{INTRODUCTION}
Autonomous flight is an open problem that has been receiving considerable attention from the robotics and machine learning research communities over the past few years with a significant push towards producing fully vision-based navigation algorithms. However, the design of such algorithms is still an open problem.
\par Event cameras have emerged, in recent years, as a promising complement to conventional frame-based cameras as vision sensors. As opposed to conventional frame cameras that capture information at every pixel at regular intervals, event cameras capture information asynchronously only at pixels where the logarithmic intensity of light changes by more than a threshold value. Thus, event cameras offer significant advantages over traditional frame-based cameras, such as complete asynchronous information capture, extremely low latency (in the order of a few microseconds), and high dynamic range ($>120dB$).
\par While there has been significant research on event-based algorithms, such algorithms for flying tasks are still rare. The creation of such algorithms is hampered by the unavailability of training data specific to flying scenarios. Furthermore, existing datasets are unable to provide the high-rate ground truths necessary for effective navigation of flying craft. The creation of real-world flying event datasets faces several challenges such as the difficulties associated with precise operation of the vehicle and synchronization overheads between various sensors, further exacerbating the lack of data. 
\par While event cameras offer many advantages aiding important navigation tasks such as optical flow - computing optical flow using only events is an ill-posed problem, as shown by researchers in \cite{lee2022fusion}. There exist scenarios where the utility of optical flow, by itself, is limited. For instance, a small, slow object closer to a camera can have the same optical flow as a larger, faster object located further away. In such scenarios, the utility of optical flow estimates can be increased by the addition of depth data. Existing datasets incorporate depth ground truth using LiDAR sensors which, though accurate, can not match the spatial and temporal resolution provided by true depth cameras. This loss of resolution can have disastrous consequences, especially in the flying scenario. Therefore, a good flying event dataset must provide accurate ground truth depth to aid effective learning.
\par Several studies \cite{lee2022fusion, gehrig2020eklt, jiang2020learning, scheerlinck2019asynchronous, pan2019bringing, pan2020single} have shown the enhanced utility of using a fusion of event data and frames for navigation tasks. Such fusion enables the extraction of visual features not provided by event cameras, such as color, along with dense temporal features from events. Such fusion is valuable in static cases, where the absence of relative motion between the event camera and the scene causes no events to be generated. Therefore, a good flying event dataset must provide high-resolution frame data to enable fused learning of tasks.
\begin{table*}[ht]
    \centering
    \def\arraystretch{1.1}%
    \setlength{\tabcolsep}{6pt} 
    \caption{Comparison of the Ground-truth provided by the relevant datasets mentioned in Section \ref{relwork}}
    \label{datcomp}
    \begin{tabular}{ccccccc}
    \hline
    \hline
        Dataset & Frame Camera & Event Camera & Depth & Optical Flow \\
        \hline
        \begin{tabular}[c]{@{}c@{}}DDD-17 \cite{ddd17} \end{tabular} & \begin{tabular}[c]{@{}c@{}} 0.1MP, monochrome,\\monocular, 10-50FPS \end{tabular} & \begin{tabular}[c]{@{}c@{}}0.1MP, monocular \end{tabular} & - & no \\  \hline
        \begin{tabular}[c]{@{}c@{}}DDD-20 \cite{ddd20} \end{tabular} & \begin{tabular}[c]{@{}c@{}}0.1MP, monochrome,\\monocular, 8-50FPS \end{tabular} & \begin{tabular}[c]{@{}c@{}}0.1MP, monocular \end{tabular} & - & no \\ \hline
        \begin{tabular}[c]{@{}c@{}}MVSEC \cite{mvsec} \end{tabular} & \begin{tabular}[c]{@{}c@{}} 0.4MP, monochrome,\\stereo baseline=10cm, 20FPS \end{tabular} & \begin{tabular}[c]{@{}c@{}}0.1MP, stereo, 10cm baseline \end{tabular} & \begin{tabular}[c]{@{}c@{}}VLP-16 Lidar\\10Hz rotation rate \end{tabular} & \begin{tabular}[c]{@{}c@{}}yes\\Freq: 10Hz \end{tabular} \\ \hline
        \begin{tabular}[c]{@{}c@{}}DSEC \cite{Gehrig21ral, Gehrig3dv2021}\end{tabular} & \begin{tabular}[c]{@{}c@{}}1.6MP, color,\\ stereo baseline=51cm, 20FPS \end{tabular} & \begin{tabular}[c]{@{}c@{}}0.3MP, stereo, 60cm baseline \end{tabular} & \begin{tabular}[c]{@{}c@{}}VLP-16 Lidar\\10Hz rotation rate \end{tabular} & \begin{tabular}[c]{@{}c@{}} yes\\Freq: 20Hz \end{tabular} \\ \hline
        \textbf{FEDORA (ours)} & \begin{tabular}[c]{@{}c@{}}\textbf{1.6MP, color,}\\\textbf{monocular, 50FPS} \end{tabular} & \begin{tabular}[c]{@{}c@{}}\textbf{0.3MP,} \textbf{monocular} \end{tabular} & \begin{tabular}[c]{@{}c@{}}\textbf{Depth Camera}\\\textbf{1.6MP, 50FPS} \end{tabular} & \begin{tabular}[c]{@{}c@{}}\textbf{yes}\\\textbf{Freq: 10, 25, 50 Hz} \end{tabular} \\
        \hline
    \end{tabular}
    \vspace{-5mm}
\end{table*}

This work aims to address the dearth of well-labeled data specific to flying tasks. 
To the best of our knowledge, FEDORA is the first fully synthetic flying event dataset offering high-rate ground truths for depth, ego-pose, and optical flow, which are crucial for navigation tasks. Importantly, FEDORA provides optical flow ground truth at three different data rates - 10Hz, 25Hz, and 50Hz, for all the frames contained in the dataset. This is a notable advancement over the current de-facto standard optical flow rate of 20Hz or lower and makes this work the first to provide ground truth for, and enable the training of, real-time optical flow. To help algorithms trained using our data to generalize better, and to enhance their applicability to scenarios encountered in the real world, we provide sequences recorded in a variety of lighting conditions, with a range of different motion, and varying environmental conditions. Hence, we believe FEDORA can spur research into novel navigation algorithms for autonomous flight operations.
\par Among existing datasets, the ones closest in scope to FEDORA are the Multivehicle Stereo Event Camera Dataset (MVSEC) \cite{mvsec} and DSEC \cite{Gehrig21ral}. MVSEC provides optical flow ground truth at 20Hz. However, the quality of this ground truth is poor, due to poor handling of occlusions and moving objects. Another shortcoming of MVSEC is the low resolution of the provided event data (0.1 megapixels). DSEC provides approximately 4000 samples of accurate optical flow ground truth. However, only a very small portion of the dataset is annotated with optical flow. Also, the ground truth rate, which is of vital importance for flying applications, is restricted to a sub-realtime value of 10Hz. 
\par The primary contributions of this work are as follows:
\begin{itemize}
\itemsep0em
    \item A Synthetic dataset (FEDORA) collected from simulations of an instrumented flying quadcopter offering accurate ground truth depth, pose, and optical flow, thus enabling the training of the entire perception pipeline on a single dataset.
    \item Higher rate ground truths compared to existing datasets, capturing high-speed motion with better accuracy, and enabling the training of real-time perception algorithms.
    \item Raw event streams, RGB frames, and calibrated Inertial Measurement Unit (IMU) data in several simulated environments with different speeds and variable illumination and wind conditions.
    \item Validation of the efficacy of FEDORA compared to existing ones using standard state-of-the-art learning models for different tasks.
\end{itemize}


\section{RELATED WORK}\label{relwork}

To the best of our knowledge, there is no dataset meant exclusively for aerial navigation tasks. All existing datasets meant for navigation tasks cater to the autonomous driving scenario. In the following paragraphs, we discuss existing driving datasets in detail.

The KITTI benchmark suite \cite{geiger2012we, menze2015object, xie2016semantic} brought standardization to the benchmarking of vision-based navigation tasks. Datasets such as the \textit{Waymo open} dataset \cite{waymo} and the \textit{nuScenes} dataset\cite{nuscenes} cater primarily to object detection and tracking tasks, and hence, do not provide ground truths for pose, events, or optical flow. Similarly, the \textit{KAIST Urban} dataset \cite{kaist} provides ground truth specific to localization and mapping. The \textit{Oxford RobotCar} dataset \cite{RobotCarDatasetIJRR} contains extensive ground truth in the form of frames, LIDAR point clouds, and pose. However, this dataset is also tailored toward localization and mapping applications. The \textit{DrivingStereo} \cite{yang2019drivingstereo} and DDAD \cite{ddad} datasets provide monocular and stereo depth ground truths respectively. Both these datasets cater specifically to depth estimation tasks.

All the datasets discussed thus far do not contain event data. This fundamentally limits their applicability to scenarios such as drone navigation, where very low response times and high-speed maneuvering are necessary. In recent years, there has been significant work on driving datasets providing event data. A few of these datasets are shown in Table \ref{datcomp}. The End-to-end \textit{DAVIS Driving Dataset 2017} (DDD17) 
\cite{ddd17} provides 12 hours of data generated using a DAVIS 346B event sensor mounted behind the windshield of a car. Included in the dataset are various auxiliary measurements from the vehicle, such as vehicle speed, steering angle, throttle position, GPS coordinates, etc. This dataset is meant for the learning of driving-related control tasks such as steering angle prediction, where latency does not impose a very harsh constraint.
The DDD20 dataset 
\cite{ddd20} is a significant update on DDD17 \cite{ddd17}, with over 51 hours of driving data logged over a variety of highway and urban settings. However, as is the case with its predecessor, DDD20 is meant for slow control tasks such as steering angle prediction.

Some of the closest analogs of our work are the Multivehicle Stereo Event Camera Dataset (MVSEC) \cite{mvsec} and DSEC \cite{Gehrig21ral}. MVSEC provides synchronized stereo events along with grayscale images, ground truth depth, optical flow, IMU readings, and pose. This data is recorded using a stereo setup of DAVIS346B event cameras and a monochrome stereo frame camera setup. Depth data is collected in the form of point clouds using a Velodyne VLP-16 LIDAR. The low resolution of the provided data, combined with the poor handling of occlusions and moving objects in the provided optical flow ground truth make this dataset unsuitable for the training of algorithms in latency-critical applications such as flight. The simulated sensors used in our work have 4 and 3 times higher frame and event camera resolutions respectively. This, coupled with our depth camera providing both higher spatial and temporal resolution as compared to the Velodyne LIDAR, and our much higher rate optical flow ground truth helps mitigate the primary limitations of MVSEC.

DSEC contains all the ground truths contained in MVSEC, albeit at significantly larger resolutions. The larger baseline values of DSEC's stereo setup also help it overcome the large depth errors introduced by MVSEC's smaller baseline. However, the larger baseline is still incapable of providing the millimeter accuracy depth ground truth required for flying applications. DSEC provides optical flow only at 10Hz, making its optical flow ground truth slower than real-time, a significant disadvantage for flying applications. The temporally dense and more accurate depth ground truth provided by our work, coupled with our faster-than-real-time optical flow ground truth helps overcome the main limitations of DSEC.

\vspace{-4mm}

\section{DATASET} \label{dataset}
Real-world driving datasets are generated by mounting the chosen sensor suite on the Dataset Collection Vehicle (DCV). Usually, a range of DCVs are used to create variety in the generated data. The DCV is then driven around in various environments and the data logged by the sensor suite is extracted at the end of each driving run. This data is then suitably post-processed, and stored in a suitable format. Creators of real-world datasets also generally provide metadata such as camera intrinsic and calibration matrices, correspondence maps between different types of data, sequence lengths, etc. Quirks associated with the data collection process or the sensor suite used to log the data are also provided.

\begin{figure}[h]
    \centering
    \includegraphics[scale=0.38]{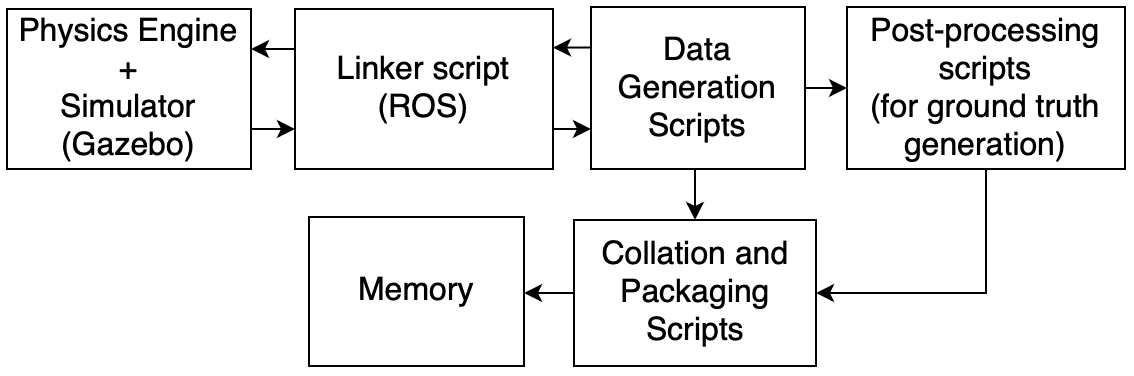}
    \caption{Worflow of the Dataset Generation Simulator}
    \label{simorg}
\end{figure}

Synthetic dataset generation environments contain a \textit{Physics Engine} that converts user-defined world and vehicle-related specifications into a simulated environment. Linker scripts link the \textit{Physics Engine} with data generation scripts running in the background. The Physics Engine controls the behavior of all objects in the simulated environment and models their interactions with each other. The DCV is instrumented with sensors that take readings from the simulated environment. These readings are post-processed to generate ``ground truths" such as optical flow. These ``ground truths" along with sensor inputs (such as camera frames) are then packaged and stored in memory. The workflow for our Dataset Generation model is as shown in Fig. \ref{simorg}.

Our dataset generation simulator leverages the open-source \textit{Physics Engine} - "Gazebo" \cite{gazebo} to simulate the vehicle dynamics and flight/motion patterns necessary to record data. We opted to use Gazebo due to its popularity in the robotics community and comprehensive documentation. 
We employ a full-fledged flight control stack alongside the data collection scripts to streamline data generation and ensure steady, level flight. This flight control stack autonomously guides the DCV along a predetermined path in simulation.
We selected the PX4\cite{px4} flight control stack utilizing the MAVLINK\cite{mavlink} communication protocol due to its lightweight nature, customizable features, and extensive developer support.
For managing the communication of sensor data generated by the simulator, we utilize the Robot Operating System (ROS)\cite{ros}, which seamlessly integrates with Gazebo and the PX4 flight stack, providing convenient access to sensor data. We use PX4's companion ground control station software - ``QGroundControl" to set the pre-planned flight trajectory. 
To eliminate data loss due to collection and processing delays, all simulations are slowed down by a factor of $32$ compared to real-time. This slowdown enables the data collection script to capture data at the high required data rates.

Multirotors are the primary aerial platforms for performing tasks targeted by this work such as optical flow and ego-motion estimation. Therefore, we opt to use a simulated quadrotor as our DCV. 
Using an easy-to-model vehicle as the DCV lowers the computational cost of the simulation while preserving its physical integrity. If such a vehicle is readily available in the real world, the ``realness" of the data collected in simulation is enhanced.
Therefore, we model our DCV on the readily available and widely used 3DR Iris quadcopter. 
The simulated sensor suite mounted on this DCV is described in the following paragraphs.

\begin{figure}[h]
    \centering
    \includegraphics[scale=0.36]{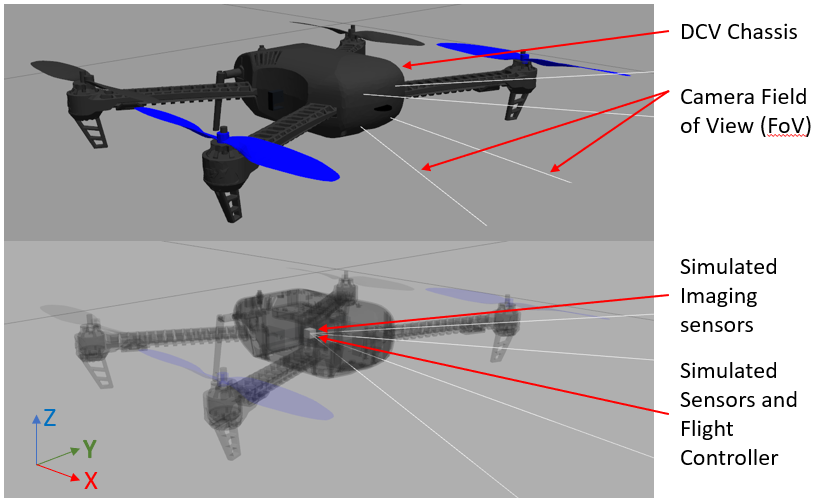}
    \caption{Simulated Quadcopter instrumented with Imaging and Inertial Sensors. The Body frame of reference of the quadcopter is shown in the bottom-left corner}
    \label{simdrone}
    \vspace{-2mm}
\end{figure}

\begin{table}[h]\centering
\def\arraystretch{1.25}%
    \caption{Specifications of the Simulated Sensors. FoV: Field of View, $fps$: Frames per second}
    \label{sensor_spec}
    \begin{tabular}{cc}
    \hline
    \hline
        Sensor & Specifications \\
        \hline
        \multirow{2}{*}{VI-sensor} & Resolution: 1440x1080 pix @50 fps \\
         & FoV: 60$^\circ$ horizontal\\
        \hline
        \multirow{3}{*}{\makecell{RealSense Depth\\sensor}} & Resolution: 1440x1080 pix @50 fps \\ & Sensing limits: [0.1, 30]m \\
         & FoV: 60$^\circ$ horizontal\\
        \hline
        \multirow{2}{*}{DAVIS} & 640x480 pixels\\
         & FoV: 60$^\circ$ horizontal\\
        \hline
        \multirow{2}{*}{IMU} & 6 DoF Inertial Data \\ & @ 1000 samples/s \\
        \hline
        \hline
    \end{tabular}
\end{table}
\par For generating RGB frames, a simulated camera based on the Gazebo Camera driver is used. The simulated camera is set up to have a focal length of $30mm$, a $60^{\circ}$ horizontal field of view, a resolution of $1440\times1080$ pixels, and a 50 fps output data rate. Each frame is timestamped and synchronized with the event and depth streams.
\par To generate events, we utilize an event camera model that can supply genuine asynchronous event streams based on the DVS plugin \cite{davis_plugin1, davis_plugin2, davis_plugin3}. Work by Gehrig, et. al.\cite{gehrighighres} suggests that a larger event camera resolution is detrimental to performance in high-speed applications such as flight. Therefore, we opt for a VGA resolution ($640\times480$ pixels) simulated event sensor. This design choice also has the desirable side-effect of making our dataset better comparable to DSEC, which also provides VGA-resolution events.
\par Due to limitations imposed by the simulator update rate, the minimum time interval between event packets is set to $0.2ms$, which while not truly asynchronous, is still found to be fast enough for the applications targetted by this work. Lower minimum time intervals may be achieved simply by further slowing down the simulation. Additionally, the event threshold of the simulated event camera is adjustable, enabling variable event densities over the same flight path. For the sequences included in this work, the event threshold is set to 30.
Recorded events are synchronized with the RGB and depth frames and the synchronization maps are provided in the dataset as metadata. This metadata contains the index of events such that $frame[i]=k$, where $frame[i]$ is the $i^{th}$ RGB or depth frame, and $k$ is the index of the earliest event such that $t_k\geq t_{frame[i]}$, where $t_n$ is the timestamp of the $n^{th}$ event/frame.

\par To generate depth, we use the Openni2\cite{openni} driver from Prime Sense. Depth frames have a resolution of $1440\times1080$ pixels and are encoded as single-precision floating-point numbers, with each pixel value representing depth in millimeters. To keep the dataset true to real-world constraints, the depth camera's range is clipped to a minimum depth of $0.1m$ and a maximum of $30m$, a constraint shared by many real-world depth cameras. Depth values beyond these thresholds are invalid and encoded by a $NaN$.
\par To ease the generation of the optical flow as described in Section \ref{gtgen}, the field of view of the RGB, depth, and event sensors is intentionally kept identical. This removes the need for numerical rectification of the Field-of-View in post-processing, as done by real-world datasets such as DSEC\cite{Gehrig21ral}.
\par Pose is recorded in the DCV's body frame using the simulator back-end and is post-processed into the global frame using the transform library ``\textit{tf}" \cite{tf}.
To better mimic real-world data characteristics, Gaussian noise is added to the recorded pose data. This gives our dataset the unique capability of being ``programmable" with respect to noise.
\begin{figure}[h]
    \centering
    \includegraphics[scale=0.3]{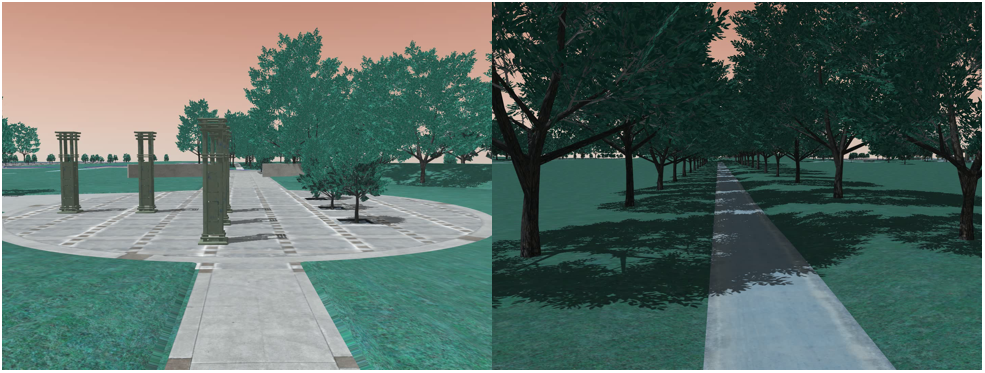}
    \caption{Sample RGB frame from sequence Baylands-Day2}
    \label{bay2}
\end{figure}
\begin{figure}[h]
    \centering
    \includegraphics[scale=0.3]{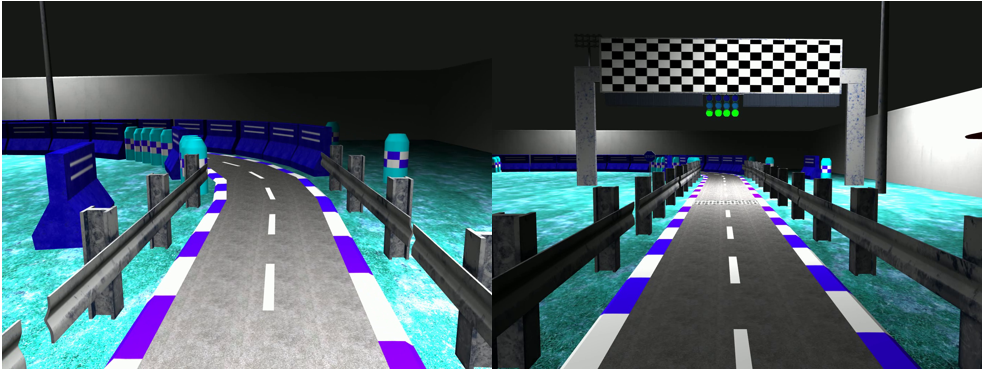}
    \caption{Sample RGB frame from sequence Racetrack-Night}
    \label{rn}
\end{figure}

\begin{figure*}[h]
    \centering
    \includegraphics[scale=0.515]{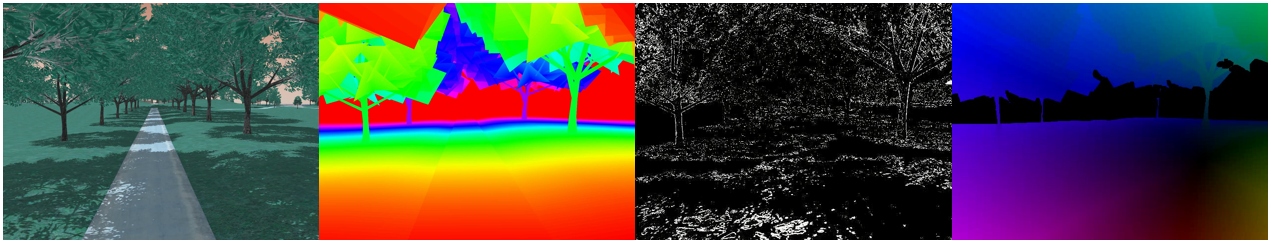}
    \caption{Example RGB and Depth data from the Baylands-Day2 sequence along with the corresponding event and optical flow renders}
    \label{sampleRender}
\end{figure*}

\begin{table}[h]
    \centering
    \caption{Sequences in FEDORA. T: Length of the sequence, D: Total distance traveled by the DCV in the simulated flight, $||v||_{max}$: Maximum aerial velocity attained by the DCV, $||\omega||_{max}$: Maximum angular velocity of the DCV in flight, MER: Mean Event Rate, MEPS: Million Events Per Second}
    \label{seq}
    \def\arraystretch{1.25}%
    \setlength{\tabcolsep}{4pt}
    \begin{tabular}{ccccccc}
    \hline
    \hline
        World & Sequence & T(s) & D(m) & \begin{tabular}[c]{@{}c@{}}$||v||_{max}$\\(m/s)\end{tabular} & \begin{tabular}[c]{@{}c@{}}$||\omega||_{max}$\\(rad/s)\end{tabular} & \begin{tabular}[c]{@{}c@{}}MER\\(MEPS)\end{tabular} \\
        \hline
        \multirow{3}{*}{Baylands} & Day1 & 165 & 594.87 & 5.032 & 1.233 & 1.12 \\
         & Day2 & 220 & 386.31 & 2.531 & 2.845 & 1.44 \\ 
         & Ngt1 & 159 & 268.47 & 2.416 & 10.718 & 1.69 \\
        \hline
        \multirow{2}{*}{Racetrack} & Day & 112 & 188.86 & 2.218 & 2.374 & 2.01 \\
         & Ngt & 112 & 188.36 & 2.058 & 2.098 & 1.79 \\
        \hline
        \hline
    \end{tabular}
\end{table} 

\subsection{Sequences}
As part of our dataset, we provide sequences from different environments such as:

\begin{enumerate}
    \item \textbf{Baylands}: This environment is a true-to-scale simulation of the Baylands Park in Sunnyvale, CA. We provide sequences from various overflights of the simulated park. Overflights are recorded in varying illumination conditions. Fig. \ref{bay2} shows a sample frame from the sequence Baylands-Day2.
    \item \textbf{Racetrack}: This is an outdoor environment from AWS 
    simulating a racetrack. We provide one daytime and night-time sequence each in this environment. Both sequences are a survey pattern of the racetrack. Fig. \ref{rn} shows an example frame from Racetrack-Ngt.
\end{enumerate}
Table \ref{seq} lists statistics related to FEDORA sequences. Fig. \ref{sampleRender} shows a sample color frame, the corresponding accumulated event frame, and depth and optical flow ground truths.


\subsection{Data Format}
\begin{figure}
    \centering
    \includegraphics[scale=0.3]{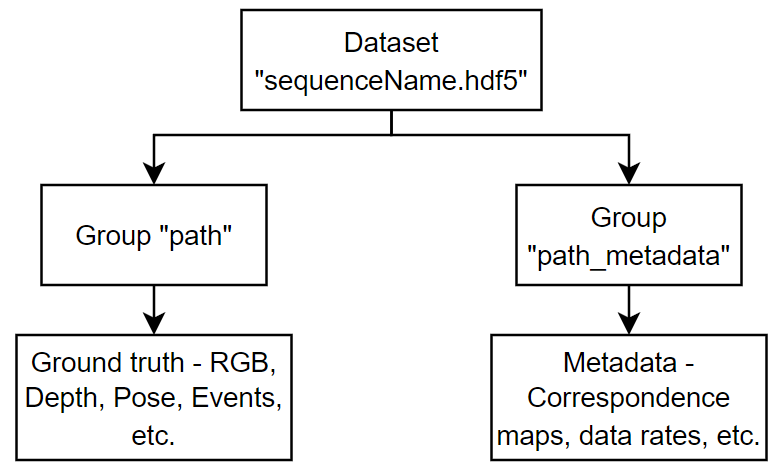}
    \caption{Dataset Storage Hierarchy}
    \label{datahier}
\end{figure}

We provide our dataset as Hierarchical Data Format, version 5 (HDF5) files, with each file corresponding to the sequences shown in Table \ref{seq}. Each file, in turn, contains a group ``path" with ground truth values for that sequence, along with a group containing metadata. Fig. \ref{datahier} diagrammatically shows this hierarchy.
The HDF5 format is chosen due to its efficient packing fraction, ability to ``lazily" load data, and widespread use and compatibility with commonly used DNN libraries such as PyTorch, TensorFlow, and Keras. 

Existing datasets are provided as multiple files spread across several directories. Such a storage architecture adds hard-to-debug failure points to downstream user scripts such as dataloaders. Using a single HDF5 file to store the dataset remedies these problems.
 
For each of the sequences in our dataset, we provide the following data packed into an HDF5 file:
\begin{itemize}
\itemsep0em 
    \item 1440x1080p RGB images at 50fps from a simulated frame camera.
    \item Inertial data from an IMU, at a rate of 1k samples/s.
    \item Events from a simulated DAVIS sensor \cite{davis_plugin1, davis_plugin2, davis_plugin3}.
    \item Depth images from the simulated depth sensor at 50fps.
    \item Pose data from the simulator processed using the ``transform library" (tf) package \cite{tf} at a rate of 200Hz.
    \item Optical flow ground-truth at three frequencies - 10, 25, and 50 Hz, generated as described in Section \ref{gtgen}.
\end{itemize}




\section{OPTICAL FLOW GENERATION}\label{gtgen}
Optical Flow is the apparent motion of a pixel on the 2D image plane due to the movement of object(s) in the observer's three-dimensional Field of View (FoV). Applications of optical flow include object detection and tracking, background separation, motion detection, and visual odometry. This makes optical flow especially useful in the flying scenario, where its ability to provide fast object detection and motion estimates enables agile drone navigation. For this reason, we choose to include the optical flow ground truth in this work. 
We generate optical flow using an analytical approach similar to the one described by \cite{Zhu_2019_CVPR}, which uses Eq. \ref{trans} and \ref{flow} to generate optical flow from the depth and pose estimates obtained from a neural network they proposed. 
\begin{equation}
    \begin{pmatrix}
    x_i^*\\y_i^*
    \end{pmatrix}
    =K\pi\bigg(R\frac{fb}{d_i}K^{-1}
    \begin{pmatrix}
    x_i\\y_i\\1
    \end{pmatrix}
    +T\bigg)
    \label{trans}
\end{equation}
\begin{equation}
    \begin{pmatrix}
    u_i\\v_i
    \end{pmatrix}
    =\frac{1}{B-1}\bigg(
    \begin{pmatrix}
    x_i^*\\y_i^*
    \end{pmatrix}-
    \begin{pmatrix}
    x_i\\y_i
    \end{pmatrix}\bigg)
    \label{flow}
\end{equation}
\begin{equation}
    K=\begin{bmatrix}
        f_x & s & x_0 \\
        0 & f_y & y_0 \\
        0 & 0 & 1
    \end{bmatrix}
    \label{intrmat}
\end{equation}
Here,
$x_i^*$ and $y_i^*$ are the new coordinates of the pixel $(x_i, y_i)$ after application of an affine pose transform which estimates the position of pixel $(x_i, y_i)$ due to inter-frame motion. This motion is described by the rotation matrix $R$ and the translation vector $T$. $K$ is the camera intrinsic matrix (shown in Eq. \ref{intrmat}), $f$ is the focal length of the camera, $b$ is the baseline of the stereo camera setup, $d_i$ is the disparity at the $i^{th}$ pixel and $B$ is the bin count. Zhu, et al. use bins to discretize the temporal dimension as part of their input representation. Thus, they compute optical flow in units of ``pix/bin" instead of the standard ``pix/s". The projection function (Eq. \ref{pifunc}), referred to as the $\pi$ function is used to normalize the depth axis, which is the $z$-axis in the coordinate frame used by Zhu, et al.
\begin{equation}
    \pi
    \begin{pmatrix}
        x & y & z
    \end{pmatrix}^T=\begin{pmatrix}
        \frac{x}{z} & \frac{y}{z}
    \end{pmatrix}^T
    \label{pifunc}
\end{equation}
In the camera intrinsic matrix, given by Eq. \ref{intrmat}, $f_x$ and $f_y$ are the focal lengths of the camera (in pixels) about the $X-$ and $Y-$axes respectively, $s$ is the skew, and $x_0$ and $y_0$ are the distances (in pixels) of the principal point of the camera (the center of the sensor) from the top-right corner of the sensor. Skew is a measure of the sheer distortion introduced into an image when the $x$ and $y$ axes of the image plane are not perpendicular to each other. In our setup, all three image sensors (RGB, depth, and event) are configured to have a skew of \textit{zero}.
\begin{figure}
    \centering
    \includegraphics[scale=0.5]{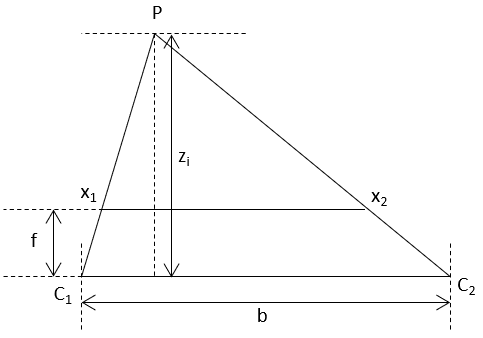}
    \caption{Depth estimation in a Stereo setup. We use Eq. \ref{simt} to modify Eq. \ref{trans} to generate optical flow}
    \label{simtri}
\end{figure}
Since we do not use a stereo setup like the one used by \cite{Zhu_2019_CVPR, mvsec}, the baseline $b$ and disparity $d_i$ cannot be computed. However, from Fig. \ref{simtri} and by the laws of similar triangles, we can derive Eq. \ref{simt}. Using Eq. \ref{simt}, the scalar term $\dfrac{fb}{d_i}$ can be replaced with the depth at the $i^{th}$ pixel, $z_i$, which is available to us as a ground truth value.
\begin{equation}
    \frac{b}{z_i}=\frac{(b+x_2)-x_1}{z_i-f}
    \implies z_i=\frac{f.b}{x_1-x_2}=\frac{f.b}{d_i}
    \label{simt}
\end{equation}

The factor $\dfrac{1}{B-1}$ represents the number of bins used to discretize the time domain. We do not use bins and compute optical flow per unit time. Thus, we can replace this factor with the inter-frame interval of the generated optical flow.

Thus, Eq. \ref{trans} and \ref{flow} can be rewritten as Eq. \ref{trans*} and \ref{flow*}.
\begin{equation}
\begin{pmatrix}
    x_i^*\\y_i^*
    \end{pmatrix}
    =K\pi\bigg(Rz_iK^{-1}
    \begin{pmatrix}
    x_i\\y_i\\1
    \end{pmatrix}
    +T\bigg)
    \label{trans*}
\end{equation}
\begin{equation}
\begin{pmatrix}
    u_i\\v_i
    \end{pmatrix}
    =t_{ie}\bigg(
    \begin{pmatrix}
    x_i^*\\y_i^*
    \end{pmatrix}-
    \begin{pmatrix}
    x_i\\y_i
    \end{pmatrix}\bigg)
    \label{flow*}
\end{equation}
where $t_{ie}$ is the interval between consecutive optical flow estimates.
Eq. \ref{trans*} holds only when the $z$ axis is the depth direction. However, as shown in Fig. \ref{simdrone}, the depth direction of our simulated DCV is the $X$ axis.
\begin{figure}[hb]
    \centering
    \includegraphics[scale=0.35]{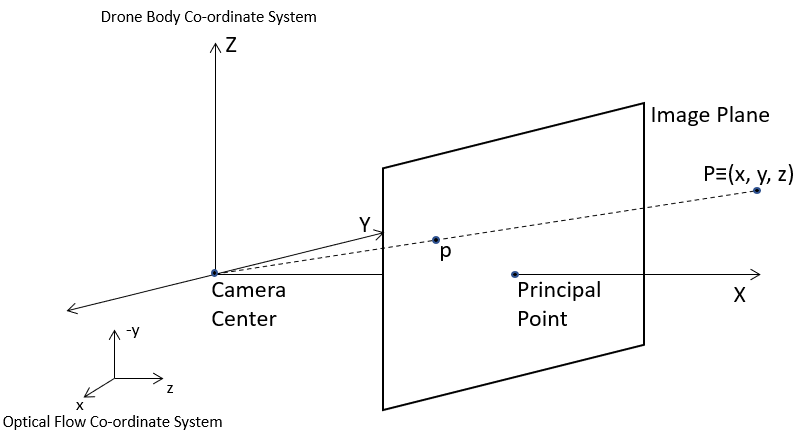}
    \caption{Co-ordinate System used by our simulated cameras and the coordinate system used for computation of Optical Flow}
    \label{camframe}
\end{figure}
To accommodate for this change, the pinhole camera model is modified as shown in Fig. \ref{camframe} This necessitates the addition of a rotation operation to the optical flow computation pipeline. To avoid introducing this additional complexity to Eq. \ref{trans*}, we apply this corrective rotation to the translation vector $T$ and rotation matrix $R$ of the drone before other computations.

As shown in \cite{Gehrig21ral}, event data must be rectified if the Fields of View (FoV) of the vision sensors are not identical. In our setup, we use a depth camera to measure depth directly. To reduce the overhead of event rectification, we design the event and depth cameras to have the same FoV and position both sensors at the same point in simulated space.

The primary advantage of our optical flow generation methodology is its programmability in terms of the rate of the generated optical flow. We provide optical flow for each of the sequences in FEDORA at three different rates - $10fps$, $25fps$, and $50fps$. However, our code may be used to compute optical flow at other frequencies too.

\section{DISCUSSION} 
High Degree-of-Freedom (DoF) depth data is required to generate 
optical flow ground truth with both high temporal resolution, and high range. Since non-flying DCVs have DoF restricted to 3, most existing datasets (DrivingStereo \cite{yang2019drivingstereo}, DDAD\cite{ddad}, and Kitti \cite{Geiger2012CVPR}) contain only 3-DoF data. Although MVSEC \cite{mvsec} and DSEC \cite{Gehrig21ral} utilize flying DCVs for some sequences, these sequences are restricted to controlled indoor environments, causing the depth data in these sequences to have little, if any components beyond the standard 3-DoF values.

The sequences in FEDORA on the other hand have been recorded in simulated with non-trivial maneuvers, thus yielding depth values along all 6-DoF. Such data is more suited to training depth estimators for flying vehicles. This, in turn, results in a higher DoF optical flow.

The mean value of optical flow provided by DSEC is $7.751$ pix/s, while the mean optical flow values for the 10 Hz optical flow ground-truth in the Baylands-Day1, -Day2, and -Ngt1 sequences in FEDORA is $18.841$ pix/s. Thus, optical flow networks trained on FEDORA are exposed to a larger range of flow values, leading to better flow estimation ability.

Another advantage of FEDORA is the high resolution of the depth provided. Some datasets such as DDAD use only LIDARs to estimate depth, thus limiting the resolution of their depth data in both the spatial and temporal domains. Others like DSEC use stereo sensing to compute depth, fundamentally limiting their spatial resolution. FEDORA uses a simulated depth camera to get depth, leading to millimeter accuracy on every valid pixel.

\section{EXPERIMENTS}\label{exp}

To assess the difference between the 6-DoF ground truth provided by our dataset and the 3-DoF ground truth common in existing datasets, we conduct a series of experiments. All experiments are performed on machines containing Nvidia A100 GPUs, and an AMD Epyc 7502 32-core processor. We provide experimental results for disparity-, pose-, and optical flow-estimation, trained on the three major outdoor sequences - Baylands-Day1, Baylands-Day2, and Baylands-Night1. All our test networks utilize the same encoder as the one proposed by \cite{mint}, differing only in their decoder stage.

\subsection{Disparity Experiments}
To examine the significance of the 6-DoF ground truth provided by this work, we train and test our test network on FEDORA and DSEC. For a fair comparison, we choose the Normalised Average Disparity Error (NADE) instead of the Average Disparity Error (ADE) as our quantification metric, computed as follows:
\begin{equation*}
    NADE=\frac{\text{ADE}}{\text{Mean Disparity Ground Truth}}
\end{equation*}
DSEC \cite{Gehrig21ral} provides disparity maps instead of the depth frames provided by our dataset.  Therefore, when training the test network on FEDORA, we convert our depth frames into disparity by computing the reciprocal of the depth value at each pixel. We expect the network to learn the associated scalar proportionality constant during training. Thus, the disparity network trained on our data must learn a more challenging problem, which explains the higher NADE value obtained by testing on our dataset, as shown in Table \ref{expReDep}.

\begin{table}[h]
\def\arraystretch{1.2}%
    \centering
    \caption{Results of training our test network on Disparity data from FEDORA and DSEC}
    \label{expReDep}
    \begin{tabular}{c c c c}
    \hline
    \hline
        Training Set & Test Set & \makecell{Mean Test Set \\ Disparity Value} & NADE \\
        \hline
        DSEC & DSEC & 20.21 & 0.1765 \\
        FEDORA & FEDORA & 0.2566 & 0.2689 \\
        \hline
        FEDORA & DSEC & 20.21 & 0.6769 \\
        DSEC & FEDORA & 0.2566 & 45.1552 \\
        \hline
    \end{tabular}
\end{table}

To verify our intuition on the differences between flying and driving datasets and the impact of higher DoF ground truth, we perform cross-testing experiments on the test network by evaluating the test network trained on DSEC on FEDORA, and vice versa. The results of these experiments, shown in table \ref{expReDep} confirm our intuition and highlight the ability of a network trained on our higher DoF, higher variance ground truth to generalize to lower DoF ground truth from DSEC.



\subsection{Optical Flow Experiments}
To assess the benefits of temporally dense optical flow, we train the modified EvFlowNet proposed by \cite{mint} on our multi-frequency optical flow ground truth. We use Normalised Average Endpoint Error (NAEE) as the result quantification metric, as the mean optical flow value differs across frequencies and datasets, making comparisons based on the Average Endpoint Error (AEE) an unfair metric. NAEE is given by 
\begin{equation*}
    \text{NAEE}=\frac{\text{AEE}}{\text{Mean Optical Flow Ground Truth}}
\end{equation*}
We train the model on $10Hz$, $25Hz$, and $50Hz$ optical flow data from all three major outdoor sequences for 100 epochs and several learning rates and choose the best results. Optical flow for all three frequencies is generated over the same time interval to enable an iso-time comparison, and hence test the intuition that higher frequency ground truth leads to better learning.
The number of training samples in the same sequence is five times greater at $50Hz$ than $10Hz$. To verify that the high NAEE at $10Hz$ is not caused solely by the smaller number of training samples, the networks are trained on $10Hz$ data from all three major sequences together, resulting in iso-sample training of all networks. The results of these experiments, as in table \ref{expRes}, show that the NAEE decreases significantly with increasing ground truth frequency, thus confirming our intuition that a higher frequency ground truth leads to better learning of a task.
\begin{table}[h]
\def\arraystretch{1.2}%
    \centering
    \caption{Results of training modified EvFlowNet \cite{mint} on data from Baylands-Day1, Baylands-Day2, and Baylands-Ngt1}
    \label{expRes}
    \begin{tabular}{c c c c c}
    \hline
    \hline
        \makecell{Optical Flow\\Frequency} & \makecell{Mean Optical\\Flow(pix/s)} & \makecell{Ground Truth\\Variance(pix/s)} & AEE & NAEE\\
        \hline
        $10$Hz & 18.841 & 18.67 & 13.44 & 0.794 \\
        $25$Hz & 10.392 & 12.163 & 5.30 & 0.51 \\
        $50$Hz & 6.15 & 9.13 & 2.448 & 0.398 \\
        \hline
    \end{tabular}
\end{table}

To validate our intuition of significant differences between driving and flying datasets, we provide results for cross-testing experiments with optical flow, where we train our test network alternatively on FEDORA and DSEC, and test on the other dataset. Since DSEC provides optical flow ground truth at 10Hz only, we perform these experiments using only 10Hz flow ground truth from FEDORA. Table \ref{expReFlow} contains the results of these experiments.\vspace{-1mm}

\begin{table}[h]
\def\arraystretch{1.2}%
\setlength{\tabcolsep}{4pt}
    \centering
    \caption{Results of training our test network on Optical Flow data from FEDORA and DSEC}
    \label{expReFlow}
    \begin{tabular}{c c c c c}
    \hline 
    \hline
        Training Set & Test Set & \makecell{Mean Test Set \\ Flow Value} & \makecell{Ground Truth\\Variance(pix/s)} & NAEE \\
        \hline
        DSEC & DSEC & 7.751 & 7.00 & 0.2211 \\
        DSEC & FEDORA & 18.841 & 18.67 & 1.346 \\
        \hline
        FEDORA & DSEC & 7.751 & 7.00 & 2.77 \\
        FEDORA & FEDORA & 18.841 & 18.67 & 0.794 \\
        \hline
    \end{tabular} 
\end{table}
The results in Table \ref{expReFlow} confirm our hypothesis on the incompatibility of sequences recorded in a fundamentally different manner. The large error observed when networks trained on FEDORA are tested on DSEC, and vice versa, is a direct result of the higher DoF data provided by this work, along with the large difference between the average optical flow between these datasets. Since FEDORA provides a significantly greater range of flow values, with a correspondingly larger variance, networks trained on it fail to estimate accurately the small range of flow values provided by DSEC. The same applies to networks trained on DSEC, which can not estimate the flow values provided by FEDORA due to their training set containing only a small range of flows. The comparatively larger error when using FEDORA as both the train and test set arises from the small size of the test network, which is optimized for the smaller range of optical flow values provided by DSEC.

\subsection{Ego-Pose Experiments}
Our dataset provides pose ground truth at 200 samples/s, which is downsampled to 50 samples/s and then used to train our test network. Since real-world driving datasets like DSEC do not provide pose ground truths, and since there is no real-world analog of our flying dataset, we only show the results of experiments on our dataset. These are as described in Table \ref{expResPose}. We use the Normalised Average Error (NAE) to quantify our results, as below:
\begin{equation*}
    NAE=\frac{\text{Average Pose Error}}{\text{Mean Pose Ground Truth}}
\end{equation*}
It can be seen from Table \ref{expResPose} that very good pose predictions can be achieved using just three FC layers after the encoder.
\begin{table}[h]
\def\arraystretch{1.2}%
    \centering
    \caption{Results of training our test network on FEDORA Pose data}
    \label{expResPose}
    \setlength{\tabcolsep}{10pt} 
    \begin{tabular}{c c c}
    \hline 
    \hline
        Ground Truth &  \makecell{Mean Test\\Ground Truth} & NAE \\
        \hline
        Orientation &  0.02698 & 0.0754 \\
        Position &  0.2566 & 0.0206 \\
    \hline
    \end{tabular}
\end{table}



\section{CONCLUSION}

We present FEDORA, a fully synthetic flying-event dataset for autonomous flight operations, containing event streams, RGB frames, and IMU data, along with depth, ego-pose, and optical flow ground truths, being the only synthetic dataset to provide all these ground truths in a single package. Notably, it provides multi-frequency optical flow ground truth, enabling both the training of real-time optical flow and the study of the trade-off between accuracy and computation.

FEDORA enables rapid efficacy checking of new model architectures, thereby speeding up the model development cycle. It also enables holistic training, leading to more accurate and energy-efficient networks in the future. We hope FEDORA will help spur research in latency-limited and safety-critical applications such as drone operations.\\


\vspace{-4mm}

\addtolength{\textheight}{-12cm}   







\bibliographystyle{IEEEtran}
\bibliography{root}


\end{document}